\documentclass[conference]{IEEEtran}
\IEEEoverridecommandlockouts
\usepackage{graphicx}
\usepackage{subfig} 
\usepackage{cite}
\usepackage{amsmath,amssymb,amsfonts}
\usepackage{algorithmic}
\usepackage{graphicx}
\usepackage{textcomp}
\usepackage{float}
\usepackage{xcolor}
\def\BibTeX{{\rm B\kern-.05em{\sc i\kern-.025em b}\kern-.08em
    T\kern-.1667em\lower.7ex\hbox{E}\kern-.125emX}}
\begin{document}

\title{Crossing Language Borders: A Pipeline for Indonesian Manhwa Translation}

\author{
    \IEEEauthorblockN{Nithyasri Narasimhan and Sagarika Singh}
    \IEEEauthorblockN{nn7985@rit.edu and ss3028@rit.edu}
}

\maketitle

\begin{abstract}
In this project, we develop a practical and efficient solution for automating the Manhwa translation from Indonesian to English. Our approach combines computer vision, text recognition, and natural language processing techniques to streamline the traditionally manual process of Manhwa(Korean comics) translation. The pipeline includes fine-tuned YOLOv5xu\cite{yolov5} for speech bubble detection, Tesseract\cite{smith_tesseract_2007} for OCR and fine-tuned MarianMT\cite{marianmt} for machine translation. By automating these steps, we aim to make Manhwa more accessible to a global audience while saving time and effort compared to manual translation methods. While most Manhwa translation efforts focus on Japanese-to-English, we focus on Indonesian-to-English translation to address the challenges of working with low-resource languages. Our model shows good results at each step and was able to translate from Indonesian to English efficiently.
\end{abstract}

\section{Introduction}
Manhwa is primarily created in their native languages, making them inaccessible to many global readers. Translating these works into other languages is a laborious process. Although Indonesian translations are more common, direct English translations are still relatively rare, leaving a significant gap for English-speaking audiences. Traditional translation methods are slow and inefficient, often requiring hours or even days to complete a single chapter. To address this, our project proposes an automated system for translating Manhwa from Indonesian to English. Using advances in computer vision and natural language processing, our goal is to make the process faster and more efficient. Our research focuses on four key tasks: detecting and extracting speech bubbles bubbles, performing Optical Character Recognition, performing Machine Translation and then finally overlaying the translated text back on Manhwa panels.

\subsection{Research Questions}
\begin{enumerate}
    \item How can existing machine learning models be adapted or enhanced to automate the translation of Manhwa?
    \item What approaches are most effective in building a robust translation system for a low-resource language like Indonesian?
\end{enumerate}

\subsection{Motivation}
The popularity of Manhwa is growing globally, but language barriers prevent many readers from enjoying these works. Manual translation methods are slow and require significant effort, making it difficult to keep up with demand. Our project aims to automate the workflow, making translations faster and accessible while fostering cross-cultural storytelling.

\subsubsection{Theoretical Motivation}
Indonesian, as a low-resource language, poses unique challenges due to limited datasets. This project contributes to NLP and computer vision advancements by addressing these gaps and exploring efficient ways to integrate image processing and language translation. 

\subsubsection{Real-World Motivation}
Automating translation benefits both the reader and the creator. Readers gain faster access to translated works, while creators and translators can save time and scale their efforts. 

\subsection{Applications}
Readers can enjoy translated Manhwa in English without long waiting periods. Translators can use this tool to speed up their workflow and increase productivity. This project contributes to advances in multi-modal machine learning, particularly in NLP and computer vision for low-resource languages.

\subsection{Example of problem and solution}
\subsubsection{Problem Statement}
A popular Manhwa is available exclusively in Indonesian, a low-resource language. International fans eagerly await an English version, but the traditional manual translation process involves several time-consuming steps. 

\subsubsection{Solution}
Our automated system addresses this problem by streamlining the entire workflow, by converting Indonesian Manhwa panels to English translated Manhwa panels. By automating these tasks, our system reduces the time required for translation from days to hours, ensuring faster and more consistent results. Furthermore, this solution demonstrates the potential of combining machine learning techniques to address the challenges posed by low-resource languages.

\section{Dataset}
The availability of datasets specific to Manhwa is quite limited. During our research, we identified a small yet relevant dataset on Roboflow. This dataset comprises 538 images, each annotated with a single class labeled as "Texts." While the dataset size is modest, it provides a foundational resource for analyzing text elements within Manhwa content.\cite{webcomics-text-selection_dataset} We divided this dataset to 465 training images and 118 validation images.\\
When considering Datasets for Machine translation from Indonesian to English, we used Identic\cite{identic} and OpenSubtitles\cite{opensubtitles}. The Identic dataset offers parallel text for translation tasks specific to low-resource language pairs, while OpenSubtitles captures informal, conversational text and slang. Together, they provide complementary strengths for developing a translation model suited for diverse text styles in manhwa. We created a smaller dataset that used elements from the both having 80\% in Training, 10\% in Validation, and 10\% in Testing. Training had over 30000 lines.

\section{Methodology}
Our pipeline automates the Manhwa translation process through these stages:
\begin{figure}[ht]
    \centering
    \includegraphics[width=0.30\textwidth]{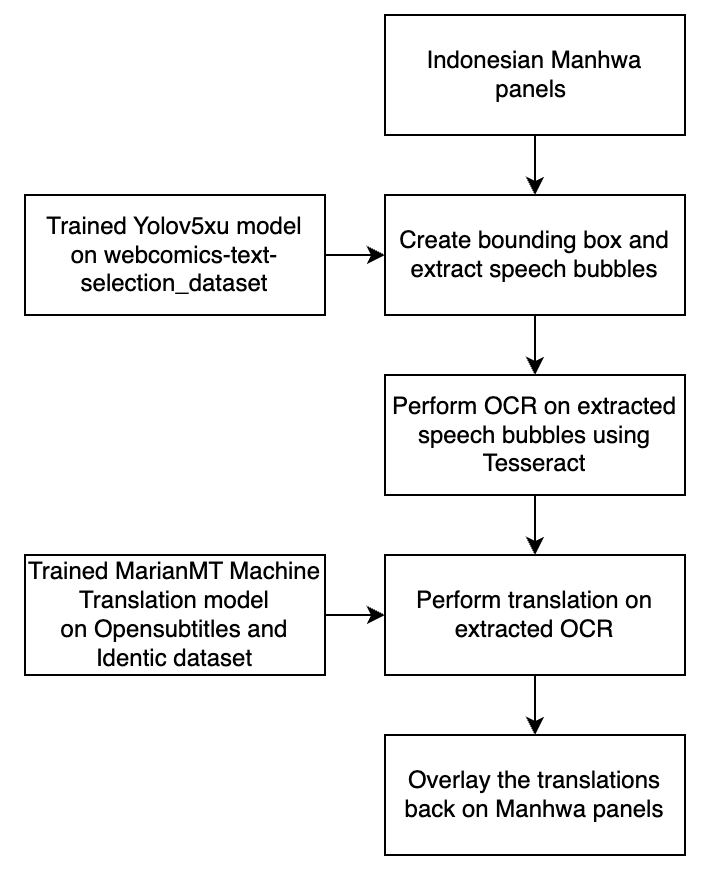} 
    \caption{Overview of Methodology}
    \label{fig:intro_image}
\end{figure}

\subsection{Speech Bubble Detection}
We used pre-trained Yolov5xu\cite{yolov5}, and fine-tuned it further on Manhwa dataset\cite{webcomics-text-selection_dataset}. YOLOv5xu is a lightweight, efficient object detection model that excels in identifying objects in images with high precision and recall. We used this fine-tuned model to detect bounding boxes of speech bubbles and extract the dimensions of the bounding boxes and extract these speech bubbles for further use. 

\subsection{Optical Character Recognition}
We performed OCR using Tesseract\cite{manga_text_recognition}, an open-source tool widely used for extracting text from images, which integrates seamlessly into our pipeline. Specifically, we utilized the Tesseract-Indonesian model with OEM set to 3 (standard) and PSM set to 6, optimized for the layout of speech bubbles in manhwa. To enhance the images, we applied grayscale preprocessing to the extracted speech bubbles. Since the images were already of satisfactory quality, no additional preprocessing steps were needed. By applying OCR directly to the extracted speech bubbles, we maximized recognition accuracy, achieving strong performance even in the presence of stylized fonts and complex layouts.

\subsection{Machine Translation}
The extracted text is translated into English using MarianMT\cite{marianmt}. MarianMT is an efficient machine translation model designed for low-resource languages. This step ensures that the translation preserves the tone, meaning, and context of the original dialogue. It was trained on the combined dataset which covered both a formal and an informal tone. This helped produce results that were able to capture the overall tone and context.

\subsection{Translated Text Overlay}
Finally, the translated text is overlaid back onto the original image. We use libraries like OpenCV\cite{scene_text_detection} and Pillow to align the new text with the original text bubble shapes, ensuring the final output looks natural and visually cohesive.

\subsection{Evaluation Metrics}
To assess the performance of our pipeline, we used the following metrics:
\begin{itemize}
    \item \textbf{Detection of Bounding Box Accuracy:} The performance of our fine-tuned Yolov5xu model for detecting bounding boxes, for speech bubbles in the Manhwa panels, was measured using Mean Precision, Mean Recall, F1 score, mAP and Mean mAP.
    \item \textbf{OCR Accuracy:} To evaluate Tesseract-OCR's \cite{smith_tesseract_2007}accuracy Character Error Rate (CER) and Word Error Rate (WER) were calculated using ground truth texts and predicted texts. CER measures the number of incorrect characters, a lower value indicates better accuracy. WER evaluates the number of incorrect words, a lower value indicates better accuracy.
    \item \textbf{Machine Translation Quality:} Evaluated using BLEU and Meteor scores to measure how well the translation captures the original meaning. BLEU measures the overlap of n-grams between the machine translation and a reference. Meteor considers word matches, synonyms, and semantic similarity.
\end{itemize}

\section{Results}

Our research questions were aimed at evaluating the adaptation of existing machine learning techniques for Manhwa translation and developing robust systems for low-resource languages like Indonesian. The following summarizes our findings and their alignment with these questions.

\subsection{Evaluation of fine-tuned Yolov5xu model}
\begin{table}[H]  
\centering  
\begin{tabular}{ |c|c|}  
  \hline
  Evaluation Metrics & Score\\
  \hline
  Mean Precision   & 89.4\%   \\
  \hline
   Mean Recall & 96.3\% \\
  \hline
  mAP@0.5 & 96.3\% \\
  \hline
  Mean mAP@0.5 & 88.9\% \\
  \hline
  F1 Score & 90.7\% \\
  \hline
\end{tabular}
\caption{Evaluation of fine-tuned Yolov5xu model}  
\end{table}
The fine-tuned YOLOv5xu model was trained on a relatively small dataset, as no large publicly available datasets exist for Manhwa text bubble detection. Despite the limited data, the model achieved an F1 score of 90.7\% demonstrating strong performance in identifying text bubbles and ensuring reliable detection. These results indicate that even with a small dataset, YOLOv5xu is capable of effectively detecting text bubbles, although the performance could likely improve with a larger, more diverse dataset.

\subsection{Evaluation of Tesseract-OCR model}
\begin{table}[H]  
\centering  
\begin{tabular}{ |c|c|}  
  \hline
    Evaluation Metrics & Score\\
  \hline
  Average Character Error Rate   & 3.1\%   \\
  \hline
   Average Word Error Rate & 8.6\% \\
  \hline
\end{tabular}
\caption{Evaluation of Tesseract-OCR model} 
\end{table}
While character-level accuracy was satisfactory, the higher word error rate suggests challenges in accurately transcribing complete words. Applying post-processing techniques might yield better results. 

\subsection{Evaluation of MarianMT-Machine Translation model}
The MarianMT translation model was fine-tuned on the OpenSubtitles and Identic datasets, which provided bilingual Indonesian-English training data. 
\begin{table}[H]  
\centering  
\begin{tabular}{ |c|c|}  
  \hline
    Evaluation Metrics & Score\\
  \hline
  BLEU   & 0.27   \\
  \hline
   Meteor & 0.61  \\
  \hline
\end{tabular}
\caption{Evaluation of MarianMT-Machine Translation model}  
\end{table}
The BLEU score reflects moderate n-gram overlap, indicating room for improvement in handling idiomatic expressions and context-specific phrases. However, the higher Meteor score suggests strong semantic retention and meaning alignment, making the model effective for this translation task, particularly in low-resource settings.

\subsection{Results of complete pipeline of our model}
Our complete pipeline successfully integrates all steps - text bubble detection, OCR transcription, machine translation, and text reintegration. Visual inspection of translated samples (e.g., Fig. 7) confirms that the pipeline preserves artistic integrity while providing accurate translations. While the final translated image is not an exact word-for-word translation, the context and meaning are not lost. Additionally, the fine-tuned YOLOv5x model demonstrated its capability to detect even small speech bubbles, ensuring comprehensive text extraction from complex layouts. Furthermore, Tesseract OCR delivered strong character-level recognition, effectively transcribing text from speech bubbles despite challenges with stylized fonts and noisy backgrounds. This demonstrates the feasibility of using our approach for automating manhwa translation tasks, even with the challenges posed by a low-resource language like Indonesian.

\begin{figure}[H]
    \centering
    \subfloat[Original Indonesian Manhwa panel]{%
        \includegraphics[width=0.20\textwidth]{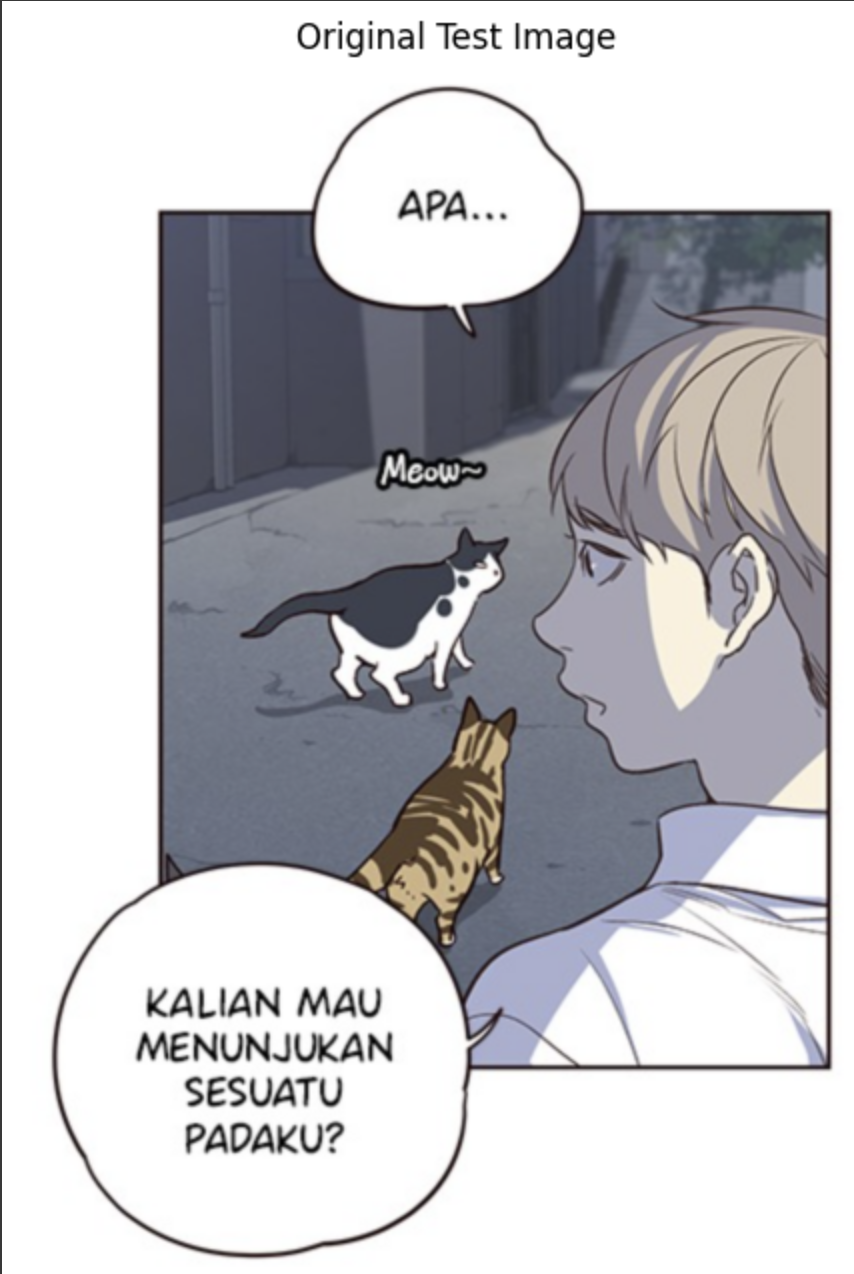}
        \label{fig:og_image}
    }%
    \hfill
    \subfloat[Detected bounding boxes for speech bubbles]{%
        \includegraphics[width=0.20\textwidth]{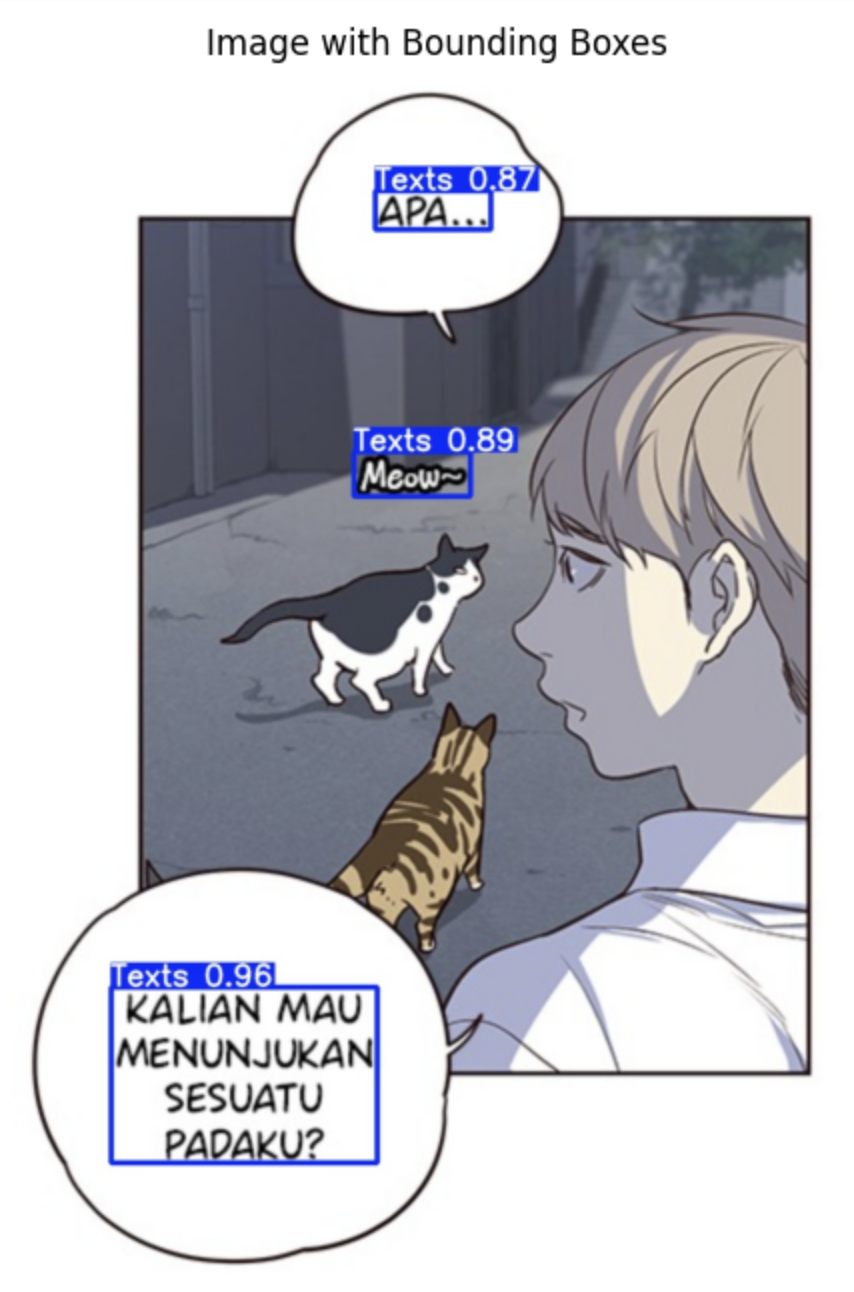}
        \label{fig:bb_image}
    }%
    \caption{Side-by-side comparison of the original panel and bounding box detection.}
    \label{fig:side_by_side}
\end{figure}

\begin{figure}[H]
    \centering
    \subfloat[Extracted enhanced speech bubbles]{%
        \includegraphics[width=0.20\textwidth]{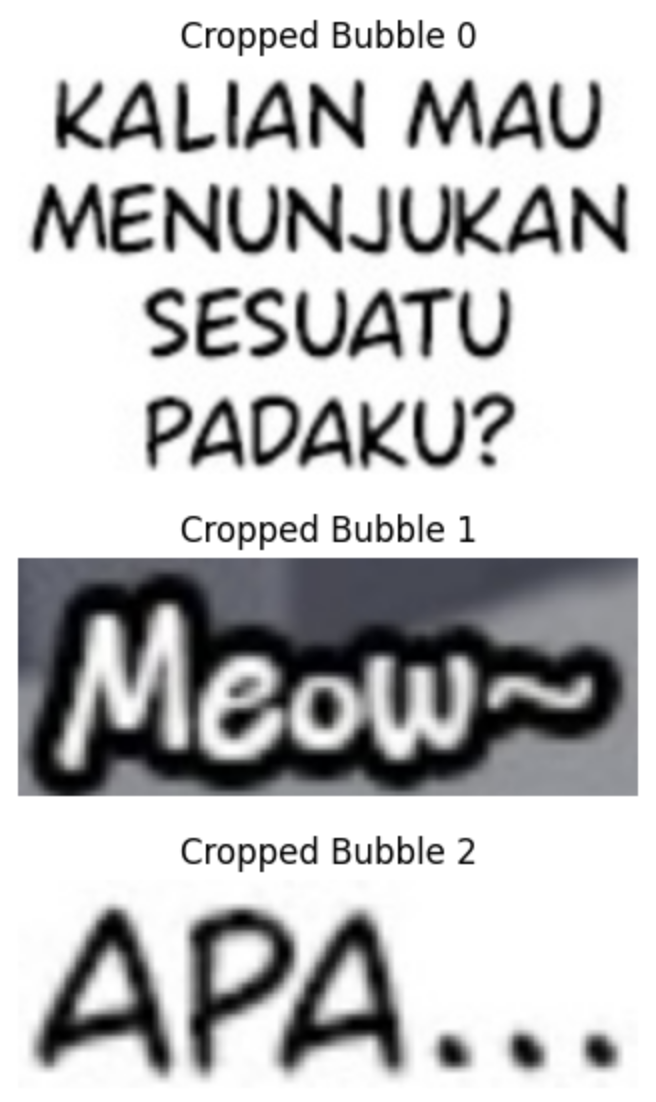}
        \label{fig:sp_b_image}
    }%
    \hfill
    \subfloat[Final translated Manhwa panel]{%
        \includegraphics[width=0.20\textwidth]{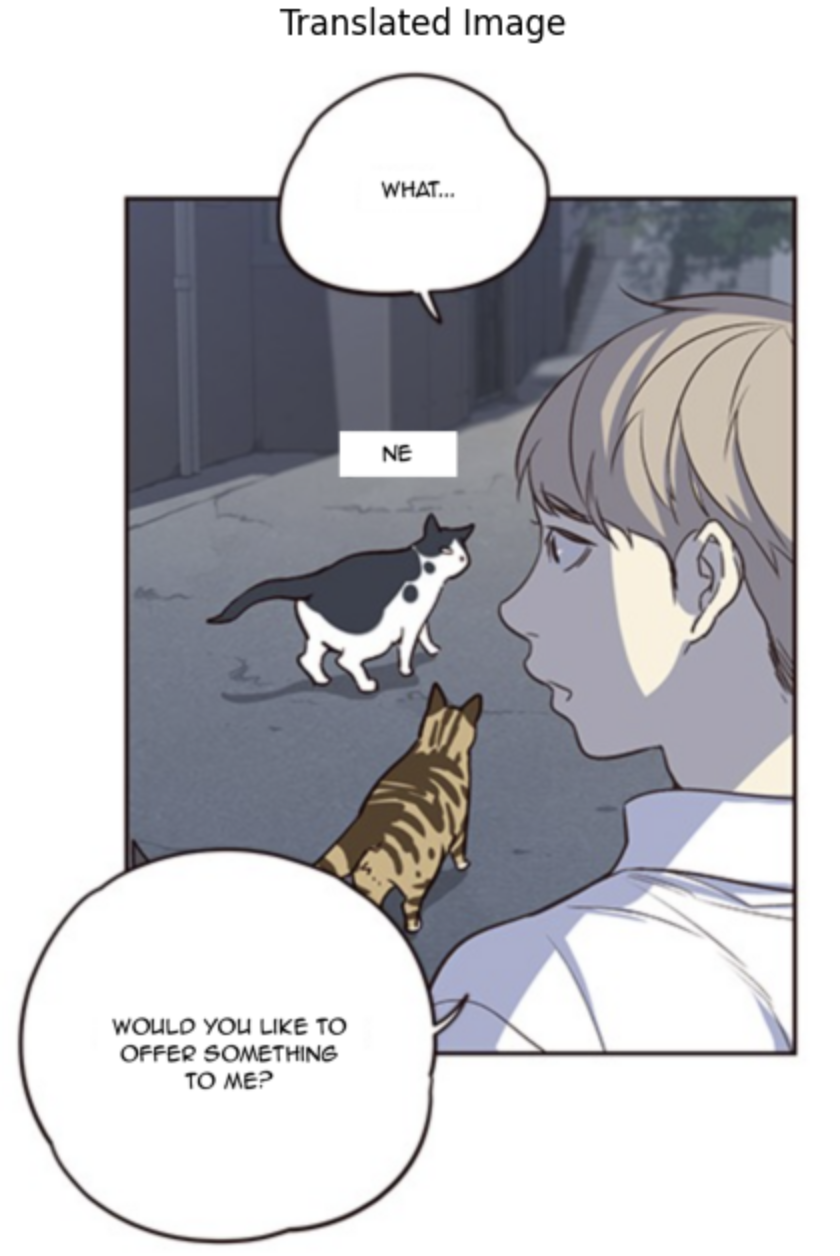}
        \label{fig:output_image}
    }%
    \hfill
    \subfloat[Results of OCR on extracted speech bubbles]{%
        \includegraphics[width=0.20\textwidth]{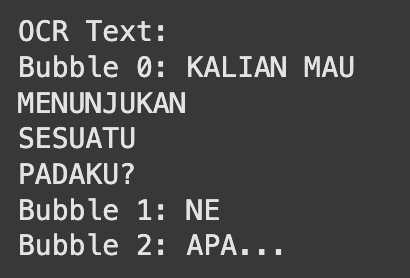}
        \label{fig:ocr_image}
    }%
    \hfill
    \subfloat[Translated text from OCR]{%
        \includegraphics[width=0.25\textwidth]{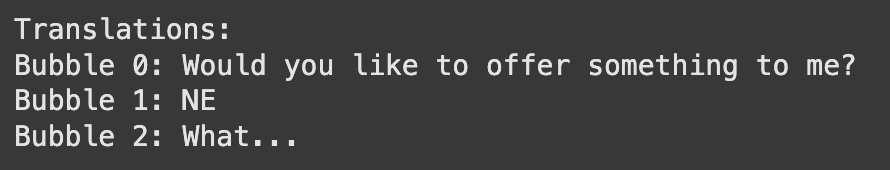}
        \label{fig:trans_image}
    }%

    \caption{Pipeline steps: (a) Enhanced speech bubbles, (b) Final translated panel, (c) OCR results, and (d) Translated text .}
    \label{fig:pipeline_steps}
\end{figure}

\section{Comparative Results}
\subsection{Object Detection Models Comparison:}
CO-DETR achieved a box mAP of 66\% on the COCO test-dev dataset for general object detection tasks\cite{inproceedings}. YOLOX-L attained a COCO-style AP of 70.2\% on the Manga109-s dataset\cite{unknown}. Our fine-tuned YOLOv5xu model achieved an mAP@0.5:0.95 of 88.9\% using the Webcomice dataset\cite{webcomics-text-selection_dataset} for text bubble detection. This result highlights the advantage of domain-specific fine-tuning and the effectiveness of using a curated dataset tailored for Manhwa.

\subsection{OCR Models Comparison:}
A segmentation-free OCR method on comic books, achieved a CER of 22.78\% and a WER of 39.30\%\cite{8270233}. Using Tesseract OCR, our pipeline achieved a CER of 3.1\% and a WER of 8.6\%, significantly outperforming the segmentation-free approach. The smaller dataset of high-quality images in our study and effective pre-processing likely contributed to these superior results.

\subsection{Machine Translation Models Comparison:}
Indonesian-English Neural Translation on IWSLT datasets gave BLEU = 23\%, and Meteor = 57\%\cite{dwiastuti-2019-english}. The MarianMT model in our pipeline achieved a BLEU score of 27\% and a Meteor score of 61\%. These scores reflect better semantic retention and formal translation quality, likely due to fine-tuning on OpenSubtitles and Identic datasets tailored for conversational text.\\

Our pipeline consistently demonstrates superior performance across all key components when compared to existing models and methodologies. The results underscore the importance of domain-specific datasets, effective pre-processing, and targeted fine-tuning. While our smaller dataset yields strong results, expanding and diversifying the data will further enhance the pipeline’s robustness and scalability for broader applications.

\section{Analysis}
The results of our pipeline demonstrate its effectiveness in automating Manhwa translation while addressing key research questions. By integrating text detection, transcription, translation, and reintegration, the pipeline significantly streamlines traditionally manual workflows. The YOLOv5xu model achieves high recall and precision, ensuring reliable detection of even small speech bubbles in complex layouts. Combined with Tesseract OCR and a fine-tuned MarianMT model, the system retains semantic accuracy and adeptly handles contextual nuances, highlighting its domain-specific focus and superior performance compared to general-purpose models.\\
While limitations persist, such as the lack of a dedicated Manhwa dataset, challenges with informal language, and computational constraints, the results validate the pipeline's potential to enhance accessibility and efficiency in Manhwa translation. This work demonstrates the power of adapting machine learning techniques to low-resource tasks, addressing both linguistic and artistic complexities, and establishing a robust foundation for future advancements in automated Manhwa translation.

\section{Conclusion}
We presented an automated system for translating Manhwa from Indonesian to English, addressing the challenges of low-resource language processing. Our pipeline effectively combines computer vision, OCR, and machine translation to streamline the traditionally manual process. Evaluation results demonstrate the strong performance of each component, with YOLOv5x achieving high detection accuracy, Tesseract handling transcription reliably, and MarianMT providing meaningful translations. The system shows potential for broader applications in translating image-based content across underrepresented language pairs.

\section{Future Work}
Future work for this project includes several directions to enhance its capabilities and impact. Expanding and annotating a larger, more diverse dataset for Manhwa is essential to improving model performance. Incorporating post-processing techniques can improve both text bubble detection and artistic consistency in text reintegration. Extending the pipeline to support additional low-resource language pairs and enabling the system to process entire chapters or multiple panels simultaneously, rather than single panels, will broaden its applicability and scalability, making it a versatile tool for global audiences.

\section{Acknowledgment}
I would like to express my gratitude to my professor, Dr. Cece Alm, for her invaluable guidance and encouragement throughout the course of this project as part of the PSYC:681-Natural Language Processing course at Rochester Institute of Technology.

\bibliographystyle{IEEEtran}
\bibliography{ref}

\begin{thebibliography}{10}
\providecommand{\url}[1]{#1}
\csname url@samestyle\endcsname
\providecommand{\newblock}{\relax}
\providecommand{\bibinfo}[2]{#2}
\providecommand{\BIBentrySTDinterwordspacing}{\spaceskip=0pt\relax}
\providecommand{\BIBentryALTinterwordstretchfactor}{4}
\providecommand{\BIBentryALTinterwordspacing}{\spaceskip=\fontdimen2\font plus
\BIBentryALTinterwordstretchfactor\fontdimen3\font minus \fontdimen4\font\relax}
\providecommand{\BIBforeignlanguage}[2]{{%
\expandafter\ifx\csname l@#1\endcsname\relax
\typeout{** WARNING: IEEEtran.bst: No hyphenation pattern has been}%
\typeout{** loaded for the language `#1'. Using the pattern for}%
\typeout{** the default language instead.}%
\else
\language=\csname l@#1\endcsname
\fi
#2}}
\providecommand{\BIBdecl}{\relax}
\BIBdecl

\bibitem{yolov5}
\BIBentryALTinterwordspacing
G.~Jocher, ``Ultralytics yolov5,'' 2020. [Online]. Available: \url{https://github.com/ultralytics/yolov5}
\BIBentrySTDinterwordspacing

\bibitem{smith_tesseract_2007}
\BIBentryALTinterwordspacing
R.~Smith, ``Tesseract {OCR} {Engine},'' 2007. [Online]. Available: \url{https://web.archive.org/web/20160819190257/tesseract-ocr.googlecode.com/files/TesseractOSCON.pdf}
\BIBentrySTDinterwordspacing

\bibitem{marianmt}
M.~Junczys-Dowmunt, R.~Grundkiewicz, T.~Dwojak, H.~Hoang, K.~Heafield, T.~Neckermann, F.~Seide, U.~Germann, A.~Fikri~Aji, N.~Bogoychev \emph{et~al.}, ``Marian: Fast neural machine translation in c++,'' in \emph{Proceedings of ACL: System Demonstrations}, 2018, pp. 116--121.

\bibitem{webcomics-text-selection_dataset}
\BIBentryALTinterwordspacing
L.~B. Nunes, ``Webcomics text selection dataset,'' \url{ https://universe.roboflow.com/luciano-bastos-nunes/webcomics-text-selection }, dec 2024, visited on 2024-12-10. [Online]. Available: \url{https://universe.roboflow.com/luciano-bastos-nunes/webcomics-text-selection}
\BIBentrySTDinterwordspacing

\bibitem{identic}
R.~M. Rosa, I.~Warsono, P.~E. Murni, W.~Ariyanti, S.~K. Ika, and M.~Adriani, ``Identic: The indonesian-english parallel corpus,'' in \emph{Proceedings of the 8th Workshop on Asian Language Resources (ALR 2010)}, 2010, pp. 71--74.

\bibitem{opensubtitles}
P.~Lison and J.~Tiedemann, ``Opensubtitles2016: Extracting large parallel corpora from movie and tv subtitles,'' in \emph{Proceedings of the 10th International Conference on Language Resources and Evaluation (LREC 2016)}, 2016, pp. 923--929.

\bibitem{manga_text_recognition}
G.~Zhang, X.~Yan, M.~Iwamura, and Y.~Matsuo, ``Automatic manga text detection and recognition,'' \emph{arXiv preprint arXiv:2012.14271v2}, 2021.

\bibitem{scene_text_detection}
Y.~Baek, B.~Lee, D.~Han, S.~Yun, and H.~Lee, ``Efficient and accurate scene text detector,'' \emph{arXiv preprint arXiv:2103.14027v3}, 2021.

\bibitem{inproceedings}
Z.~Zong, G.~Song, and Y.~Liu, ``Detrs with collaborative hybrid assignments training,'' 10 2023, pp. 6725--6735.

\bibitem{unknown}
Y.~Shinya, ``Usb: Universal-scale object detection benchmark,'' 03 2021.

\bibitem{8270233}
C.~Rigaud, J.-C. Burie, and J.-M. Ogier, ``Segmentation-free speech text recognition for comic books,'' in \emph{2017 14th IAPR International Conference on Document Analysis and Recognition (ICDAR)}, vol.~03, 2017, pp. 29--34.

\bibitem{dwiastuti-2019-english}
\BIBentryALTinterwordspacing
M.~Dwiastuti, ``{E}nglish-{I}ndonesian neural machine translation for spoken language domains,'' in \emph{Proceedings of the 57th Annual Meeting of the Association for Computational Linguistics: Student Research Workshop}, F.~Alva-Manchego, E.~Choi, and D.~Khashabi, Eds.\hskip 1em plus 0.5em minus 0.4em\relax Florence, Italy: Association for Computational Linguistics, Jul. 2019, pp. 309--314. [Online]. Available: \url{https://aclanthology.org/P19-2043}
\BIBentrySTDinterwordspacing

\end{thebibliography}

\end{document}